\documentclass{IOS-Book-Article}

\usepackage{amssymb}
\usepackage{amsmath}
\usepackage{graphicx}

\usepackage{color}

\newcommand{\wrt}{{\it w.r.t. }}   
\newcommand{\eg}{\emph{e.g.}, }       
\newcommand{\ie}{\emph{i.e.}, }      
\newcommand{\etc}{\emph{etc. }}

\usepackage{amsthm}

\theoremstyle{definition}
\newtheorem{definition}{Definition}

\theoremstyle{definition}
\newtheorem{property}{Property}

\def\hb{\hbox to 10.7 cm{}}

\begin{document}

\pagestyle{headings}
\def\thepage{}

\begin{frontmatter}              

\title{A Visual Distance for WordNet}

\markboth{}{April 2018\hb}

\author[A]{\fnms{Raquel} \snm{P\'erez-Arnal}}, %
\author[A]{\fnms{Armand} \snm{Vilalta}}, %
\author[A]{\fnms{Dario} \snm{Garcia-Gasulla}},
\author[B]{\fnms{Ulises} \snm{Cort\'es}},
\author[B]{\fnms{Eduard} \snm{Ayguad\'{e}}}
and
\author[B]{\fnms{Jesus} \snm{Labarta}}
\address[A]{Barcelona Supercomputing Center (BSC)\\ (\{raquel.perez, armand.vilalta, dario.garcia\}@bsc.es)}
\address[B]{Barcelona Supercomputing Center (BSC) \\ Universitat Polit\`ecnica de Catalunya - BarcelonaTech (UPC) }

\begin{abstract}


Measuring the distance between concepts is an important field of study of Natural Language Processing, as it can be used to improve tasks related to the interpretation of those same concepts. WordNet, which includes a wide variety of concepts associated with words (\ie synsets), is often used as a source for computing those distances. In this paper, we explore a distance for WordNet synsets based on visual features, instead of lexical ones. For this purpose, we extract the graphic features generated within a deep convolutional neural networks trained with ImageNet and use those features to generate a representative of each synset. Based on those representatives, we define a distance measure of synsets, which complements the traditional lexical distances. Finally, we propose some experiments to evaluate its performance and compare it with the current state-of-the-art.
\end{abstract}

\end{frontmatter}
\markboth{April 2018\hb}{April 2018\hb}

%
%
%
%
%
%

\section{Introduction}
To compare two lexically expressed concepts is a problem that pervades much of Natural Language Processing (NLP) applications, such word sense disambiguation, question answering, determining the structure of texts, information extraction and retrieval, \etc  
A lot of measures had been proposed for this purpose \cite{Budanitsky2006,Lin1998, Resnik1995,Wu1994}, mostly based on distances defined on the WordNet \cite{Fellbaum1998} hierarchy.

WordNet \cite{Fellbaum1998} is a well known lexical database composed of nouns, verbs adjectives and adverbs interlinked using conceptual, semantic and lexical relations. Each concept is represented by a synset, which includes all synonymous word meanings. In the bibliography, there are several distances proposed, all of them based on lexical semantic properties of the synsets to compare.

In contrast with those previous works, we propose a new approach, defining a VD based on the visual similarity of two WordNet synsets. For setting this distance, we use the ImageNet dataset, which contains a large number of images mapped to WordNet synsets. Such a VD offers a new perspective to the previously mentioned tasks and could be used to extend and enrich the information given by the other existing distances through the consideration of a new modality (\ie images). 

To define a visual distance, we first generate a representation of WordNet synsets based on visual features. This is done through the use of pre-trained convolutional neural networks (CNN). CNN are representation learning models, capable of learning patterns of relevance for the characterisation of data. By capturing and extracting the neural activations produced within them, we generate a representation of synsets in the exceptionally rich language learnt by a deep CNN.

Recent studies show that features learned by a CNN for a given purpose can be reused for a different task \cite{yosinski2014transferable, he2016deep, Garcia-Gasulla2017a}. This illustrates the potential generality of features learned by these models. Furthermore, results from \cite{Garcia-Gasulla2017a, Garcia-Gasulla2017b} indicate that images belonging to the same synset will share a significant amount of features. From these observations, we make a further hypothesis regarding the generalist nature of CNN features and propose a formal mathematical distance. Through this distance, we may characterise, for example, WordNet synsets, and extract information on their inter-relations. We do so by first creating a representative of each synset, and then applying our distance measure.


\section{Related Work}\label{sec:SOA}


Concept similarities and distances have been previously explored mostly through the use of lexical data. WordNet, as a large-scale lexical database, is often used for this purpose, mainly through the analysis of its hyponym/hypernym taxonomy \cite{Wu1994,Leacock1998}. Other lexical based approaches are based on computing the information content of words, which in turn is obtained through statistics of their use in a given corpus \cite{Resnik1995,Jiang1997,Lin1998}. These types of approaches are extensively discussed and compared in \cite{Budanitsky2006}. 


In this work, we propose a similarity and distance between concepts based on the visual properties found in images tagged to the corresponding WordNet synset. This multimodal approach (\ie pictures and text) has been recently explored within the neural network community, through the definition of multimodal embeddings. Simply put, multimodal embeddings generate two representations, one for text and one for images, and fit those two representations into a single, shared embedding space \cite{Vilalta2017,Dong2016}. Although these approaches have shown great power at bridging both modalities (\eg for image captioning or image retrieval), their interpretability is severely limited.

In parallel, the field of transfer learning focuses on the reuse of a representation language, learned by a neural network model to solve one particular problem, to solve a different problem. In essence, by processing images through a pre-trained network and capturing the intermediate neural activations, one can obtain the \textit{perception} of that image according to the pre-trained net. When using a large dataset for training (both regarding classes and in instances per class), such representation language can be extremely rich and descriptive. In this paper we follow this approach, using such neural activations as the source for defining a visual distance of WordNet concepts. We use a deep neural representation based on the full-network embedding.




\subsection{Full-Network Embedding} \label{sec:FNE}

The Full-Network Embedding (FNE) \cite{Garcia-Gasulla2017b} is an embedding that integrates convolutional and fully connected layers from a CNN and generates a representation space of the data instances passed through. The embedding is generated from the neural activations of those layers. Activations which go through some post-processing steps until getting a discretised value of either {-1,0,1}, representing their relevance for characterising an instance.

 The first step to generate the FNE is to forward pass each data instance of the new dataset through the pre-trained model, capturing all the internal activations of the full network. 
    
Each filter within a convolutional layer generates several activations for a given input, as a result of convolving the filter throughout the input. To avoid a significant increase of the dimensionality, and to avoid the redundancy of filter information, it is performed an average spatial pooling on each convolutional filter,
such that a single value per filter is obtained by averaging all its spatially-depending activations. 
The values resulting from the spatial pooling are concatenated with the features from the fully connected layers into a single vector, to generate a complete image embedding (for deep CNN up to tens of thousands of floating point values).

Then it is needed a feature standardisation to put the value of each feature in the context of the dataset, as the features are obtained from neurons of different type and location (this implies significant variations in the feature activations). 
This process transforms each feature value so that it indicates how separated the value is from the feature mean regarding positive/negative standard deviations.

\begin{figure}[h!]
\centering
\includegraphics[width=\textwidth]{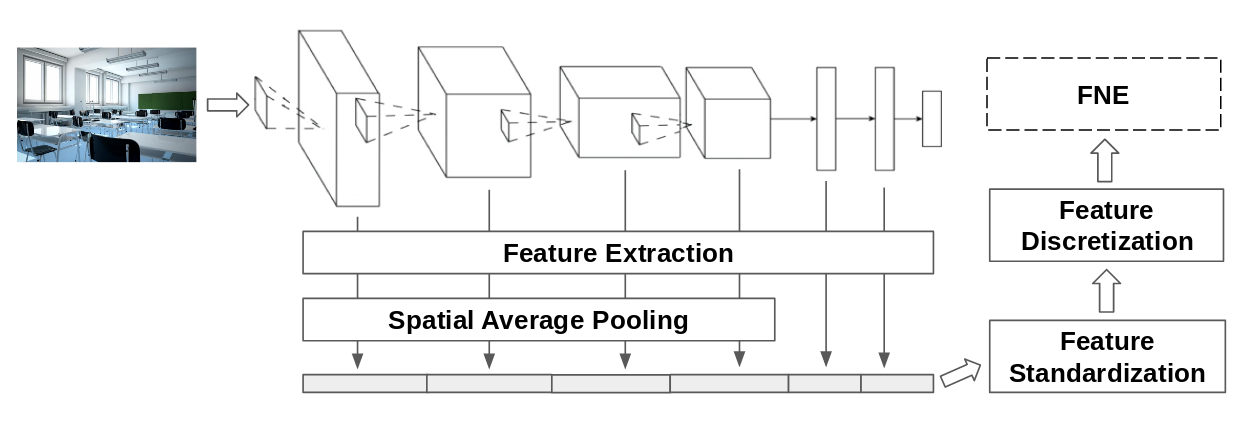}
\caption{ Generation workflow of the full-network embedding. Adapted from  Garcia-Gasulla \textit{et al}. \cite{Garcia-Gasulla2017b}. }
\label{fig:fne}
\end{figure}

Finally, it is performed a feature discretisation, to reduce the expressiveness of the features. 
This discretisation is done by mapping all the feature values to the $\{ -1, 0, 1\}$ domain by defining two thresholds $ft^−$ and $ft^+$. As shown in \cite{Garcia-Gasulla2017a}, the optimal values of these thresholds can be found empirically. This way it is represented if the feature has an atypically high value (1), typical value (0) and an atypically low value ($-1$). Authors name the features with value $-1$ \textit{characteristic by absence}, with value $0$ \textit{uncharacteristic}, and with value $1$ \textit{characteristic by presence}. Therefore, this mapping not only reduces the dimension of the space, but it also has conceptual value. Figure \ref{fig:fne} shows a diagram of the process used to obtain an FNE.

\section{Preliminary Analysis}
 \label{sec:preliminary}
 
Towards the definition of a distance between concepts based on visual properties, we first performed a statistical analysis on the behaviour of FNE image representations for certain WordNet synsets \cite{Perez2018}. Such study included the frequency and distribution of the features of each type, in the full embedding and the different layers, the correlation between the specificity of a synset with its proportion and distribution of the features, and the relations between synsets of the same semantic families and the embeddings related to them.  
 

This study was performed using 50,000 images (validation set of the ImageNet 2012 challenge dataset), belonging to 1,000 different classes. Some of the main conclusions obtained were:

\begin{enumerate}
\item The proportion of features characteristic by absence (approx. 68\%) is significantly superior to the proportion of the features characteristic by the presence (approx. 23\%) or not characteristic (approx. 9\%).
This proportion was different between the two types of layers (convolutional and fully connected). But inside each type, the ratios were very similar. 

\item  A sample of synsets showed to have the same proportion of features of each type. 
Furthermore, the co-occurrence of features characteristic by presence increases with the specificity of the synset, making the representations of those synsets more consistent \wrt the images that compose it.


 
\item Synsets from the same semantic family (\eg dog and mammal) shared more features characteristic by presence than synsets from different semantic families (\eg goat and car).


\end{enumerate}

Considering the observations of the preliminary analysis and previous studies about the behaviour of the features learned by a CNN \cite{Garcia-Gasulla2017a}, we expect that those properties could be generalised to a dataset with more classes than the ones used to train the original net.


\subsection{Hypothesis} \label{sec:hypothesis}
The properties found in \ref{sec:preliminary} are specific of the FNE calculated for the experiments, as it was generated using a specific dataset, but it is expected that can be generalised to different embeddings. 
Based on this we make the next hypothesis: 

\begin{enumerate}
\item \label{h:discriminate} Features characteristic by the presence, those which have an abnormally high activation, could be used to discriminate all WordNet synsets. 

\item \label{h:synsetspecific} The co-occurrence of features characteristic by presence increases with the specificity of the synset (\ie with its position within the hypernym/hyponym taxonomy).

\item \label{h:features} Similar synsets (in a visual semantic sense, not necessarily in a WordNet taxonomy sense), have a higher number of shared features which are characteristic by presence.

\end{enumerate}


\section{Formalization} \label{sec:procedure}

\begin{figure}[h!]
\centering
\includegraphics[width=\textwidth]{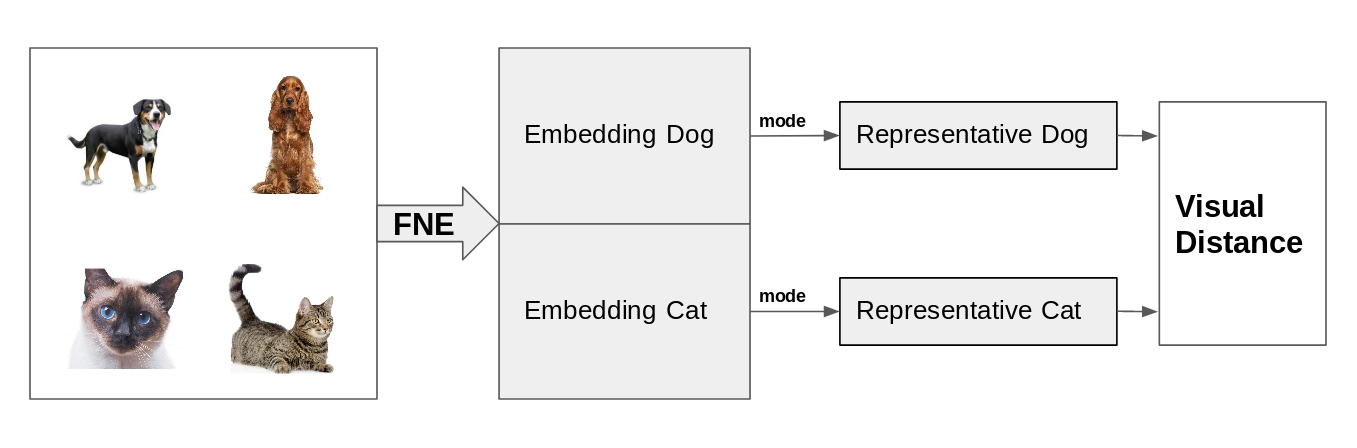}
\caption{ Generation workflow of the visual distance.}
\label{fig:visualdistance}
\end{figure}

Based on the results of \S\ref{sec:preliminary}, we will formalise the definition of the visual distance. The method used is based on, first extract the graphical features of the images from a CNN using an FNE, obtaining this way an embedding for every image. Then generate a representative from every synset. And finally, compare these representatives using the visual distance. We show this process  in Figure \ref{fig:visualdistance}. All the method tries to take advantage of the hypothesis from \S\ref{sec:preliminary} to generate a distance as most representative as possible.

\subsection{Notation}

In the next sections, we will use the following notation. 
We call the FNE of a synset $\mathbf E_{synset}$. This is a matrix of $N\times M$  where $N$ is the number of samples, and $M$ is the number of features extracted by the FNE. The number of features depends on the original network architecture and is the same independently on how many samples you have, \eg, an FNE generated over VGG16 \cite{Simonyan2014} has $M = 12,416$ and $N$ variable depending on the number of images used to produce the embedding. We also will use the notation $\mathbf E_{synset}[i,j]$ to refer to the value of the $i$ sample and the $j$ feature.

\subsection{Synset Representatives} \label{subsec:representationeuclidean}

Since there can be many samples for every synset if we use the embeddings as given it would require sizeable computational resource both regarding memory and computing power. And in the case where the two synsets we want to compare had a different number of samples, it could be affected by the differences of the sizes. The simplest way to solve these problems is to generate a representative vector for every full embedding, \ie for every synset. 
We will call the representative of a synset $\mathbf r_{synset}$, and its elements $\mathbf r_{synset}[i]$ where $i$ is between 0 and the maximum number of elements, that we will call $n_{synset}$. 
The vector is defined based on the next mathematical properties: 

\begin{enumerate}
\item All the vectors have to maintain the domain of the single embedding of samples.  
 
 \begin{equation}
      \mathbf r_{synset}[i] \in \{-1,0,1 \} \ \ \ \forall i \in [0,n]
 \end{equation}

\item All the representative vectors need to have the size of the original embedding. To not lose features in this process and have all the representatives in the same space. 
\begin{equation}
    |\mathbf r_{synset}| = n_{synset} = M \ \ \ \forall synset \in \mathbb{S} 
\end{equation}

\item A change in the images used should not significantly change the representative.
\end{enumerate}  

Our proposed option is to use the mode for every feature. It will give us a vector of the length of the number of features with values in the set $\{-1,0,1 \}$, independently of the number of samples. The number of samples needs to be big enough so that the mode gives a stable value which is usually the case in Imagenet. 

\begin{equation}
    \mathbf r_{synset}[i] = mode_{j}(\mathbf E[i,j])
\end{equation}

\subsection{Distance Formalisation}

Based on the representatives introduced in the previous section and the hypothesis of \S\ref{sec:hypothesis}, we now formalise a definition of distance, with the goal of applying it to the representatives of WordNet synsets. There are specific desirable properties on such distance: to differentiate the synsets, to be bounded, and being a mathematical metric. 

The main idea used to define the visual distance is to measure the changes between the representatives of the two synsets. Specifically,  the changes of the features characteristic by the presence, since according to the Hypothesis \ref{h:discriminate}, it is expected that this type of features will differentiate better the synsets than the not characteristic or the characteristic by absence.

\begin{definition}
    
    Given two synsets, $s_1$ and $s_2$ and it's correspondent representatives ($r_{s_1}$ and $r_{s_2}$). And given the set of possible feature index $F = [0,1,...,M]$, we define the next values.  
    
    \begin{itemize}
        \item We define $F_{(1,j)}(s_1,s_2) \subseteq F$ as the set of features that have value 1 in $r_{s_1}$ and and value $j$ in $r_{s_2}$. 
        \item We define $F_{(i,1)}(s_1,s_2) \subseteq F$ as the set of features that have value $i$ in $r_{s_1}$ and value 1 in $r_{s_2}$. 
    \end{itemize}

    \begin{align}
    F_{(1,j)}(s_1,s_2) &= \{ k \in F :  r_{s_1}[k] = 1, \ \ r_{s_2}[k] = j \ \} \\
    F_{(i,1)}(s_1,s_2) &= \{ k \in F :  r_{s_1}[k] = i, \ \ r_{s_2}[k] = 1 \ \} 
    \end{align}
    We will call $C_{(i,j)}(s_1,s_2)$  to the number of features that have value $i$ in the first representative and value $j$ in the second one and either $i$ or $j$ are 1.
    \begin{equation}
        C_{(i,j)} =  \begin{cases}
         |F_{(1,j)}| & \mbox{if } i = 1 \\ 
         |F_{(i,1)}| & \mbox{if } j = 1 \\ 
         |F_{(1,1)}| & \mbox{if } i = 1, j =1 \\ 
    \end{cases}
    \end{equation}

    Where $i$ and $j$ can be values of the set $\{-1,0,1\}$.

    
    
\end{definition}

    The visual distance is based on the value $C_{(1,1)}$ \ie characteristic by presence features found on either of two representatives. By looking at this value, we may have a initial sense of similarity between the two corresponding synsets.
    To formalise a visual distance, we normalise this value by the total number of characteristic by the presence features found on the two representatives. 
     
     The purpose of this scaling is to circumvent the variance among representatives, caused by the specificity of synsets. Since more specific synsets have a larger amount of characteristic by presence features in their representatives (as introduced in Hypothesis \ref{h:synsetspecific}), scaling by the total number of features with such value makes the distance measure invariant to synset specificity.

\begin{definition}
\label{def:similarity}
    We define the \textit{visual similarity} as the proportion of the sum common features characteristic by presence of the representatives ($C_{(1,1)}(s_1,s_2)$) respect to the sum of the $C_{(i,j)}(s_1,s_2)$.
    \begin{equation}
    sim(s_1,s_2) = \frac{C_{(1,1)}(s_1,s_2)}{\left( C_{(1,-1)} + C_{(1,0)} + C_{(1,1)} + C_{(0,1)} + C_{(-1,1)}\right)(s_1,s_2)}
\end{equation}
\end{definition}

\begin{definition}
    \label{def:distance}
     Finally we define the \textit{Visual Distance} (VD) related to the visual similarity:
     \begin{equation}
         vd(s_1,s_2) = 1 - sim(s_1,s_2)
     \end{equation}
\end{definition}

Where $s_i$ are the representatives of the synsets, we want to compare. 

\subsection{Properties of the Visual Distance}


Based on the definition, the VD has some mathematical properties.

\begin{property}
\textit{The VD is a pseudo-metric. }

The VD is generated upon the visual similarity, the definition of which is a proportion (see Definition \ref{def:similarity}). Accordingly, the distance will be symmetric and provide values within the range $[0,1]$. Thus, the VD is always non-negative and symmetric. 

Also, it is easy to prove using induction over the length of the representative that fulfils the triangle inequality. It does not satisfy \textit{the identity of indiscernibles} (Two elements have distance zero between them if and only if they are the same element). As it only depends on the features characteristic by presence, there can be some synsets that have distance zero but different representatives. This is unlikely to happen, as the representatives can have many features. In case it does happen, it would mean that the images from the two synsets are \textit{very} similar, for example, the synsets \textit{goat} and \textit{mountain goat}. 

\end{property}


\begin{property}
\textit{ The possible values given by the metric are within the range $[0,1]$.}

That is because the definition of the distance is based on a proportion so that it will be bounded. This property can help compare this distance with other distances and makes the value given by the metric more human understandable.

\end{property}

\section{Experiments}\label{sec:Experiments}

To validate the hypothesis defined in \S\ref{sec:hypothesis}, we propose the following experiments, which are left for future work.

\subsection{Generation and Study of the Distance Matrix}\label{sec:distancematrix}

First, it is proposed to generate the VD matrix of all the synsets of ImageNet. The main idea of getting this matrix is to find the different relations that could contain. 
Also, we propose to use it to perform a clustering analysis. As it is explained in \S\ref{sec:preliminary}, it is expected that synsets from the same semantic family tend to be more similar between them, regarding the VD, than compared with different families. For example, the hyponyms of \textit{mammals} like dog, goat or cat were more similar between them than compared with the hyponyms of \textit{objects} like \textit{car} or \textit{boat}. It is expected to find similar relations in the full distance matrix. 

Finally, an MDS or a PCA 2-dimensional visual representation of the full distance matrix and sections of them could give additional information about the VD.

\subsection{Comparison with Distances from WordNet}

We can compare the VD with some of the lexical and semantic ones defined in \cite{Budanitsky2006}, using the distance matrix proposed in \S\ref{sec:distancematrix}. Specifically, we aim to compare it with the path similarity distance, the Wu and Palmer similarity \cite{Wu1994} and the Lin similarity \cite{Lin1998}, as all three are metrics and return values in the same domain and codomain. 
The proposed methods to compare the distances are: 

\begin{itemize}
\item Use the Pearson or Spearman correlation coefficient to compare the different distance matrix.  
\item Compare the 2-dimensional MDS or PCA visualisation of the full matrix of the VD and the matrix of the lexical distances. Also, to compare the visualisation of some specific sections, for example, families of hyponyms like living \textit{things} or \textit{mammals}.  
\item Compare the clusters found using each one of the distance matrix. Is not expected that the visual and the lexical distances share clusters, but it could give new information about the relations between the differences or similitude of the distances. 
\end{itemize}


\section*{Acknowledgements}
This work is partially supported by the Joint Study Agreement no. W156463 under the IBM/BSC Deep Learning Center agreement, by the Spanish Government through Programa Severo Ochoa (SEV-2015-0493), by the Spanish Ministry of Science and Technology through TIN2015-65316-P project, and by the Generalitat de Catalunya (contracts 2014-SGR-1051).

\bibliographystyle{unsrt}
\bibliography{biblio}

\end{document}